\documentclass[final,a4paper,11pt]{article}

\usepackage{fancyheadings}
\usepackage{graphicx}
\usepackage{epic}

\usepackage{algorithmic}
\usepackage[]{algorithm}

\renewcommand{\baselinestretch}{1.5}

  \footrulewidth 0pt

\providecommand{\vnstitle}[1]{\title{\Large\textbf{#1}}}

\providecommand{\vnsauthor}[3]{\index{{#2, #1}}#1 #2#3}

\providecommand{\vnsinstitution}[3]{\renewcommand{\baselinestretch}{1.25}\small #1#2 \\\texttt{#3} \\ \vspace{0.2cm}}

\providecommand{\institutions}[1]{\date{#1}}

\begin{document}

\vnstitle{Variable Neighborhood Search for the University
Lecturer-Student Assignment Problem}

\author{
    \vnsauthor{Martin Josef}{Geiger},
    \vnsauthor{Wolf}{Wenger}
}

\institutions{
\vnsinstitution{}
    {Lehrstuhl f\"{u}r Industriebetriebslehre,\\Universit\"{a}t Hohenheim,\\D-70593 Stuttgart, Germany}
    {\{mjgeiger,w-wenger\}@uni-hohenheim.de}
}

\maketitle

\begin{abstract} {\em
The paper presents a study of local search heuristics in general
and variable neighborhood search in particular for the resolution
of an assignment problem studied in the practical work of
universities. Here, students have to be assigned to scientific
topics which are proposed and supported by members of staff. The
problem involves the optimization under given preferences of
students which may be expressed when applying for certain topics.

It is possible to observe that variable neighborhood search leads
to superior results for the tested problem instances. One instance
is taken from an actual case, while others have been generated
based on the real world data to support the analysis with a deeper
analysis.

An extension of the problem has been formulated by integrating a
second objective function that simultaneously balances the
workload of the members of staff while maximizing utility of the
students. The algorithmic approach has been prototypically
implemented in a computer system. One important aspect in this
context is the application of the research work to problems of
other scientific institutions, and therefore the provision of
decision support functionalities.}
\end{abstract}

{\bf Keywords:} Variable Neighborhood Search, University
Lecturer-Student Assignment Problem, Multi Objective Optimization

\pagebreak

\section{Introduction}
An integral part of a successful completion of most university
studies is the participation in seminars within which a written
report on a given scientific subject has to be produced and handed
in by a student. The topics are usually proposed by members of
staff that also support the work carried out by the students. For
many departments the assignment of students to topics and
therefore to lecturers is a recurring task that usually has to be
carried out once per term. In practice, the assignment is done
with respect to a set of aspects combining preferences of
students/lecturers as well as balancing the workload of staff.

Although the precise situation studied in the work presented here
has not been tackled in literature before, similar problems are
available in the context of \emph{course assignment} and
\emph{faculty assignment}. The general problem there lies in
assigning teachers to courses with respect to a set of side
constraints and personal preferences \cite{yang:1989:article}.
Most quantitative formulations are based on integer linear
programming models \cite{breslaw:1976:article} and solved using
commercially available software packages
\cite{saber:2001:article}. Other resolution methods used are based
on genetic algorithms \cite{wang:2002:article} or model explicit
knowledge taken from human experts
\cite{partovi:1995:article} in order to solve the problem.

Problems in the mentioned area have already quite early been
considered to employ multiple criteria \cite{mcclure:1987:article}
as preferences of a set of persons have to be met at once and
therefore require the identification of a compromise solution.
While goal programming here plays an important role, another
methodology that has been successfully applied in the past is the
Analytic Hierarchy Process AHP \cite{ozdemir:2004:article}.

The paper is organized as follows. In the following section
\ref{sec:problem}, the problem of assigning students to
topics/lecturers is explained and formulated as studied in the
real world case at the University of Hohenheim. Section
\ref{sec:experiments} presents a study of local search approaches
used to solve the problem at hand. Experiments are carried out on
a set of problem instances, one of them taken from the real world
situation and others generated on the basis of the practical case.
Conclusions are derived in section \ref{sec:conclusions}.

\section{\label{sec:problem}The university lecturer-student assignment problem}
The studied problem of assigning students to lecturers has to be
solved by the Department Production and Operations Management of
the University of Hohenheim on a recurring basis. Within a
seminar, organized twice a year, each member of staff proposes a
number of scientific topics for which students may apply.
Depending on the number of applications, identical topics may be
assigned to different students to increase the capacity of the
seminar. It can be experienced that the acceptance of a particular
topic and the corresponding motivation to work in it depends on
the personal preferences students may have for particular
subjects.

In the past the assignment of students to topics has been done on
a first-come-first-served basis, leading to a potential
unsatisfying situation for some late enrolling participants.
Students have already before been given the possibility of
articulating their preferences by ranking topics, but when
assigning topics priority was given with respect to the sequence
in which the applications have been received.

The concept currently under investigation allows each student $i$
to articulate his/her preferences by assigning nonnegative weights
$w_{ij}$ to each topic $j$ which correspond to a measure of
utility associated with the assignment of a certain topic. For
practical reasons, a total weight of $\sum_{j} w_{ij} = w_{max}$
has been chosen and communicated to the applicants, a potential
alternative however would be the normalization of the weight
values. Applications are collected up to a certain deadline, and
an optimization problem of maximizing the realized utility values
is formulated and solved as follows.

\begin{equation}
\label{eqn:zielfunktion}\max \,\,\mbox{UTILITY}  = \sum_{i=1}^{n} \sum_{j=1}^{m} w_{ij} \,\, x_{ij}\\
\end{equation}
subject to
\begin{eqnarray}
\label{eqn:minimum}&& \sum_{i=1}^{n} x_{ij} \geq a_{j}  \quad \forall j=1, \ldots, m\\ %
\label{eqn:maximum}&& \sum_{i=1}^{n} x_{ij} \leq b_{j}  \quad \forall j=1, \ldots, m\\ %
\label{eqn:alle:zuordnen}&& \sum_{j=1}^{m} x_{ij} = 1  \quad \forall i=1, \ldots, n\\ %
\label{eqn:wertebereich}&& x_{ij} \in \{0,1\}  \quad \forall i=1, \ldots, n, \,\, j=1, \ldots, m %
\end{eqnarray}

Expression (\ref{eqn:zielfunktion}) maximizes the total utility
associated with an assignment $x_{ij}$ of students $i$ to topics
$j$. The assignment is done with respect to some side constraints,
namely the minimum number $a_{j}$ and maximum number $b_{j}$ of
students per topics given in expressions (\ref{eqn:minimum}) and
(\ref{eqn:maximum}), and (\ref{eqn:alle:zuordnen}) and
(\ref{eqn:wertebereich}) make sure that every student is assigned
to exactly one topic. In the practical case, values of $a_{j} =
\left\lfloor \frac{n}{m} \right\rfloor \forall j=1,\ldots, m$ and
$b_{j} = \left\lceil \frac{n}{m} \right\rceil \forall j=1,\ldots,
m $ have been chosen for $n$ students applying for $m$ topics.

In addition to the preferences articulated by the students, some
applicants sometimes express the wish to be assigned together as a
team to a certain topic. The current procedure of defining an
assignment allows this, but only if such an alternative is also
optimal with respect to expression (\ref{eqn:zielfunktion}).

\section{\label{sec:experiments}Experimental investigation}
\subsection{\label{sec:neighborhood:search}Neighborhood search}
Although most practical instances of the model described in
section \ref{sec:problem} may be solved using e.\,g. CPLEX, a
drawback lies in the identification of a single optimal
alternative only. In order to allow an identification of an
optimal alternative that bears certain characteristics as the
mentioned simultaneous assignment of students to a particular
topic, a procedure that identifies a whole set of equally optimal
alternatives is necessary. We therefore investigated the
effectiveness of local search techniques for the problem at hand.
Starting from an initial feasible alternative, applicable
neighborhoods to the problem are in detail:
\begin{itemize}
    \item \lq{}swap2\rq{} neighborhood $nh_{swap2}$, randomly selecting two
    variables $x_{ij}, x_{kl} \mid j \neq l \wedge x_{ij} = x_{kl} = 1$ and changing the values
    to $x_{ij} = x_{kl} = 0, x_{il} = x_{kj} = 1$.
    \item \lq{}swap3\rq{} neighborhood $nh_{swap3}$, randomly selecting three
    variables $x_{ij}, x_{kl}, x_{op} \mid j \neq l \wedge l \neq
    p \wedge p \neq j \wedge x_{ij} = x_{kl} = x_{op} = 1$ and changing the
    values to $x_{ij} = x_{kl} = x_{op} = 0, x_{il} = x_{kp} = x_{oj} = 1$.
    \item \lq{}shift\rq{} neighborhood $nh_{shift}$, randomly selecting a
    variable $x_{ij} \mid x_{ij} = 1 \wedge \sum_{k=1}^{n} x_{kj} >
    a_{j}$ and a topic $l \mid \sum_{k=1}^{n} x_{kl} < b_{j}$ and
    changing the values to $x_{ij} = 0, x_{il} = 1$.
    \item \lq{}shift+swap2\rq{} neighborhood $nh_{shift+swap2}$, combining the two
    neighborhoods $nh_{shift}$ and $nh_{swap2}$ by applying both changes to the
    variables.
\end{itemize}

In addition to the application of a single neighborhood only, a
variable neighborhood search concept that includes all
neighborhoods in a set $\mathcal{NH}$ is used to solve the
problem. The general idea is sketched in algorithm \ref{alg:VNS}.

\begin{algorithm}[!ht]%
\caption{\label{alg:VNS}Reduced variable neighborhood search
(VNS)}
\begin{algorithmic}[1]%
\STATE Generate initial solution $S$, set $P^{approx} = \{ S \}$%
\REPEAT%
    \STATE Randomly select some $S \in P^{approx}$%
    \STATE Randomly select some neighborhood $nh_{i}$ from the set of neighborhoods $\mathcal{NH} = \{ nh_{1}, \ldots, nh_{q} \}$%
    \STATE Compute neighboring solution $S'$ by applying $nh_{i}$ to $S$%
    \STATE Update $P^{approx}$ with $S'$%
\UNTIL{termination criterion is met}%
\STATE Return $P^{approx}$%
\end{algorithmic}%
\end{algorithm}%

In contrast to established methods of variable neighborhood search
\cite{hansen:2003:incollection}, algorithm \ref{alg:VNS} maintains
a set $P^{approx}$ containing alternatives of equal quality with
the aim of allowing the identification of a whole set of optimal
solutions.

\subsection{Maximizing the total utility}
Experiments have been carried out for a real world problem
instance \lq{}\texttt{n34}\rq{} with $n = 34$ students applying
for $m = 15$ topics. In order to study the behavior of the
neighborhoods when facing problems with differing characteristics,
instances from $n = 30$ to $n = 45$ students have been generated
by randomly discarding or adding students from the instance
\texttt{n34}.

Each neighborhood has been tested on each problem instance
starting from an initial randomly generated solution, and average
values of the total utility as given in expression
(\ref{eqn:zielfunktion}) have been computed over a total of 25
test runs for each problem instance/neighborhood structure
combination. The termination criterion has been set to 100,000
evaluations.  On an Intel Pentium IV processor running at 3 GHz
each test run takes less than 3 seconds. The running time is
independent from the size of the problem instance as evaluation of
alternatives is possible in constant time.

\begin{table}[!ht]\centering
\begin{tabular}{lrrrrr}
Instance &  $nh_{swap2}$ & $nh_{swap3}$  & $nh_{shift}$  & $nh_{shift+swap2}$ & VNS\\
\hline
\texttt{n30}&238,34&241,73&n/a&n/a&240,99\\
\texttt{n31}&248,36&252,32&213,72&253,44&255,52\\
\texttt{n32}&257,49&262,21&239,48&265,74&268,05\\
\texttt{n33}&263,16&268,80&252,16&273,68&275,00\\
\texttt{n34} $\star$ &271,15&273,80&269,50&279,88&281,68\\
\texttt{n35}&278,88&280,96&280,28&288,76&289,92\\
\texttt{n36}&289,80&292,64&290,48&298,68&299,84\\
\texttt{n37}&298,36&301,79&300,77&308,76&309,94\\
\texttt{n38}&308,12&309,84&309,88&318,52&318,84\\
\texttt{n39}&317,04&320,36&318,40&328,12&328,84\\
\texttt{n40}&325,72&328,44&325,85&336,30&336,90\\
\texttt{n41}&333,48&335,36&331,52&343,24&344,20\\
\texttt{n42}&340,56&344,00&337,00&350,64&352,32\\
\texttt{n43}&350,16&355,44&344,56&358,36&361,24\\
\texttt{n44}&357,76&362,08&323,96&356,20&365,24\\
\texttt{n45}&365,25&368,92&n/a&n/a&369,15\\
\hline
\end{tabular}
\caption{\label{tbl:results}Average results for the UTILITY}
\end{table}

The average results of the total utility are given in table
\ref{tbl:results} with the real world instance \texttt{n34} marked
$\star$. It can be seen that VNS outperforms the simple local
search approaches in almost all problem instances with the only
exception being \texttt{n30}. For the problem instance taken from
the real world case, an optimal solution is known with a total
utility of 282, and the VNS successfully identifies one or several
optimal alternatives in almost every test run. The single
neighborhood with the best results is consistently
$nh_{shift+swap2}$, and while the results are close to the ones of
VNS, they still remain inferior.

\begin{figure}[!ht]
\centerline{\includegraphics[width=11cm]{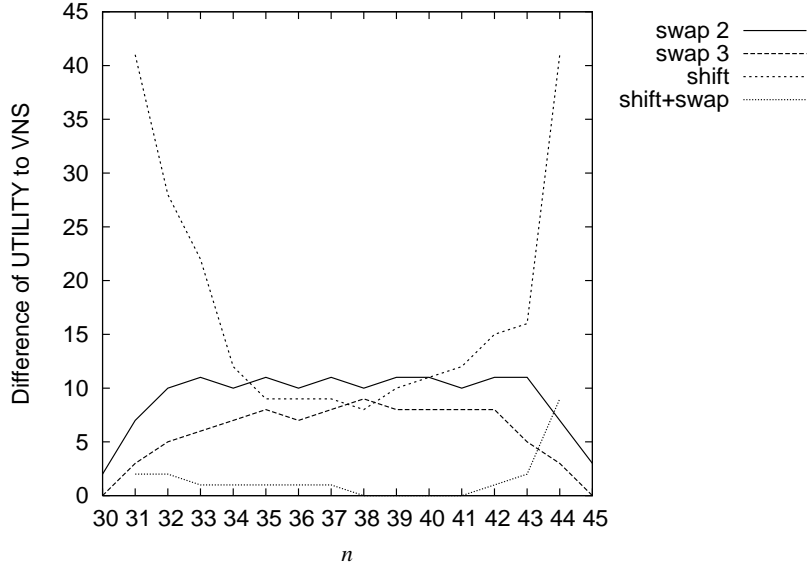}}
\caption{\label{fig:abstaende}Difference of the average UTILITY
to the results achieved by VNS}
\end{figure}

An interesting resolution pattern can be recognized depending on
the structure of the problem instances. Figure \ref{fig:abstaende}
plots and analyzes the difference of the results achieved by
simple local search to the results of VNS. The gap in performance
clearly shows that instances with little possibilities to shift
assignments are not successfully solved with $nh_{shift}$. On the
other hand, the relative performance of the swap neighborhoods
$nh_{swap2}$ and $nh_{swap3}$ decreases with increasing potential
of applying shift moves.

\subsection{Balancing utility and distribution of workload}
The encouraging results of the VNS including the neighborhoods
described in section \ref{sec:neighborhood:search} led to the
application of the search concept to a multi objective extension
of the problem. While the first objective function, given in
expression \ref{eqn:zielfunktion}, maximizes the total utility of
the students, assignments may lead to an imbalanced workload for
the members of staff as some topics tend to be significantly more
popular than others and therefore receive higher weight
assignments $w_{ij}$.

In the studied case of the real world problem, four members of
staff $\mathcal{B} = \{B_{1}, \ldots, B_{4} \}$ are involved, with
three persons supporting three topics each ($B_{1} = \{1, 2, 3\}$,
$B_{1} = \{4, 5, 6\}$, $B_{1} = \{7, 8, 9\}$) and one person being
responsible for six topics ($B_{4} = \{10, \ldots, 15\}$). In
order to balance the assignment of students, the difference
between the lecturer with the most load and the member of staff
with the lowest load is minimized. The load is here expressed as
the ratio between the number of assigned students to the number of
available spaces of the specific lecturer, given in expression
(\ref{eqn:zielfunktion2}). The problem is then solved by
identifying all optimal alternatives in the sense of
Pareto-optimality.

\begin{equation}
\label{eqn:zielfunktion2} \min \,\, \mbox{IMBALANCE} = \left(
\max_{B_{k} \in \mathcal{B}} \left( \frac{\sum_{i=1}^{n} \sum_{j
\in B_{k}} x_{ij}}{\sum_{j \in B_{k}} b_{j}} \right) - \min_{B_{k}
\in \mathcal{B}} \left( \frac{\sum_{i=1}^{n} \sum_{j \in B_{k}}
x_{ij}}{\sum_{j \in B_{k}} b_{j}} \right) \right)
\end{equation}

\begin{figure}[!ht]
\centerline{\includegraphics[width=11cm]{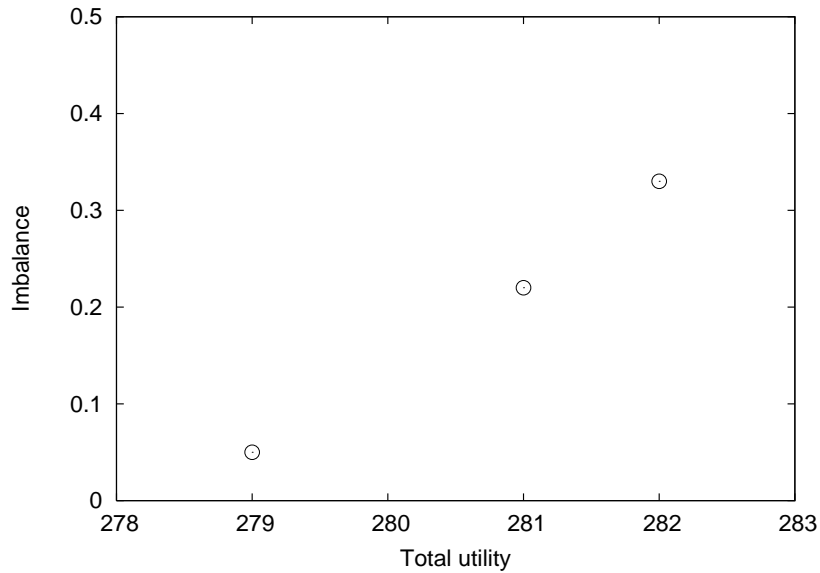}}
\caption{\label{fig:2ziele}Efficient frontier of \texttt{n34} with
optimality criteria (\ref{eqn:zielfunktion}) and
(\ref{eqn:zielfunktion2}).}
\end{figure}

The efficient solutions are plotted in outcome space in figure
\ref{fig:2ziele}. It can be shown that in addition to the optimal
alternatives for the optimality criterion of maximizing the total
utility, other efficient points exist that are better balanced
while being slightly worse in terms of their utility. A total of
139 distinct alternatives possessing values of UTILITY = 279 and
IMBALANCE = 0.055 have been identified, 29 alternatives for the
outcome of 281/0.222, and 13 for the values 282/0.333.

\section{\label{sec:conclusions}Conclusions and further research}
The paper presented a study of a real world assignment problem,
consisting of the assignment of student to scientific topics and
therefore members of staff. The problem has to be solved on a
recurring basis, and the implications of the assignments are of
high practical and personal value to the parties involved.

A quantitative model to describe the situation at hand has been
formulated and solved using a local search heuristic. Variable
neighborhood search led to superior results, independent from the
underlying problem characteristics. In addition to the
maximization of the student's utility, a second criterion has been
introduced balancing the workload of the lecturers. Pareto optimal
alternatives could have been identified, showing a tradeoff
between imbalance and utility in outcome space.

The study conducted demonstrates the applicability of local search
heuristics in general and variable neighborhood search in
particular, but is so far only based on the experiences gathered
at the University of Hohenheim. To generalize the conclusions, we
are currently evaluating the status quo of the assignment process
within seminars in universities throughout Germany with the final
goal of proposing a more general methodology that supports the
problem found in practice. The investigation is carried out
involving a questionnaire and personal communication with staff
being involved in organizing seminars.

\begin{figure}[!ht]
\centerline{\includegraphics[width=11cm]{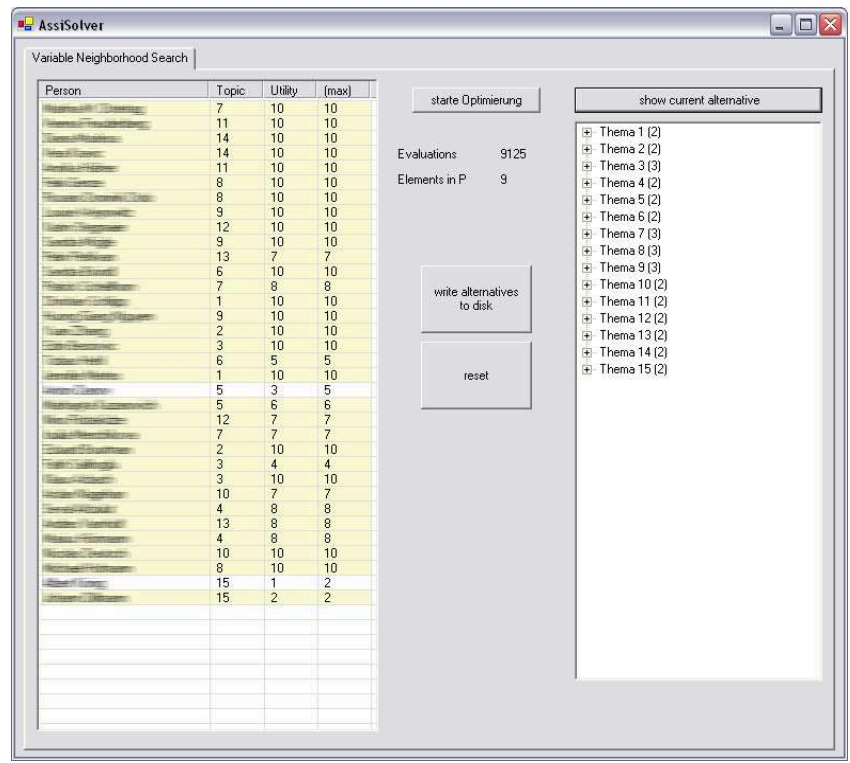}}
\caption{\label{fig:user:interface}User interface of the software}
\end{figure}

To allow the application of the optimization procedure to real
world problems, a prototypical implementation within a computer
system has been made available. Figure \ref{fig:user:interface}
shows the current user interface with which the user may interact
and which visualizes the alternatives computed by the system. The
names of the students are however here blurred due to privacy
restrictions. As the problem also has to be solved by institutions
with a low affinity to the use of computer programs, an integrated
system providing decision support may increase acceptance in
practice.

\section*{Acknowledgements}
The authors would like to thank PD Dr Andreas Kleine for providing
an optimal alternative for the real world instance \texttt{n34} of
the data sets using CPLEX.

\bibliography{../../lit_bank,../../lit_bank_nv}
\bibliographystyle{plain}

\end{document}